\documentclass[letterpaper, 10 pt, conference]{ieeeconf}  

\IEEEoverridecommandlockouts                              

\usepackage{amsmath} 
\usepackage{amssymb}  
\usepackage{subfigure}
\usepackage{siunitx}
\usepackage{booktabs}
\usepackage{color}

\input{abkuerzung.sty}
\usepackage[pdftex]{graphicx}
\usepackage{multirow}
\usepackage{url} 
\usepackage{thmtools}
\declaretheorem[name=Def., 
refname={Def.,Defs},
Refname={Def., Defs}]{Theorem}
\usepackage{hyperref}

\graphicspath{{../pdf/}{../jpeg/}}
\DeclareGraphicsExtensions{.pdf,.jpeg,.png}

\usepackage{fancyhdr}
\renewcommand{\baselinestretch}{0.92}

\title{\LARGE \bf
	ISO/TS 15066: How Different Interpretations Affect Risk Assessment}

\author{Robin Jeanne Kirschner, Nico Mansfeld, Saeed Abdolshah and Sami Haddadin$^{}$
\thanks{}
\thanks{$^{}$All authors are with Institute for Robotics and System Intelligence, Munich School of Robotics and Machine Intelligence, Technical University of Munich, 80797 Munich, Germany
	{\tt\small robin-jeanne.kirschner@tum.de}}%
}
\definecolor{babyblue}{rgb}{0.54, 0.81, 0.94}
\definecolor{orange}{rgb}{1.0, 0.65, 0.0}
\newcommand{\tabitem}{~~\llap{\textbullet}~~}

\begin{document}

\maketitle
\thispagestyle{fancy} 
\pagestyle{fancy} 

\begin{abstract}
The current technical specification ISO/TS 15066:2016(E) for safe human-robot interaction contains logically conflicting definitions for the contact between human and robot. This may result in different interpretations for the contact classification and thus no unique outcome can be expected, which may even cause a risk to the human. In previous work, we showed a first set of implications. This paper addresses the possible interpretations of a collision scenario 
as a result of the varying interpretations for a risk assessment. With an experiment including four commercially available robot systems we demonstrate the procedure of the risk assessment following the different interpretations of the TS. The results indicate possible incorrect use of the technical specification, which we believe needs to be resolved in future revisions. For this, we suggest tools in form of a \emph{decision tree} and \emph{constrained collision force maps}, which enable a simple, unambiguous risk assessment for HRI. 
\end{abstract}



\section{INTRODUCTION}

\begin{figure}[t]
	\centering
\begingroup%
  \makeatletter%
  \providecommand\color[2][]{%
    \errmessage{(Inkscape) Color is used for the text in Inkscape, but the package 'color.sty' is not loaded}%
    \renewcommand\color[2][]{}%
  }%
  \providecommand\transparent[1]{%
    \errmessage{(Inkscape) Transparency is used (non-zero) for the text in Inkscape, but the package 'transparent.sty' is not loaded}%
    \renewcommand\transparent[1]{}%
  }%
  \providecommand\rotatebox[2]{#2}%
  \newcommand*\fsize{\dimexpr\f@size pt\relax}%
  \newcommand*\lineheight[1]{\fontsize{\fsize}{#1\fsize}\selectfont}%
  \ifx\svgwidth\undefined%
    \setlength{\unitlength}{229.01675727bp}%
    \ifx\svgscale\undefined%
      \relax%
    \else%
      \setlength{\unitlength}{\unitlength * \real{\svgscale}}%
    \fi%
  \else%
    \setlength{\unitlength}{\svgwidth}%
  \fi%
  \global\let\svgwidth\undefined%
  \global\let\svgscale\undefined%
  \makeatother%
  \begin{picture}(1,0.38881168)%
    \lineheight{1}%
    \setlength\tabcolsep{0pt}%
    \put(0,0){\includegraphics[width=\unitlength,page=1]{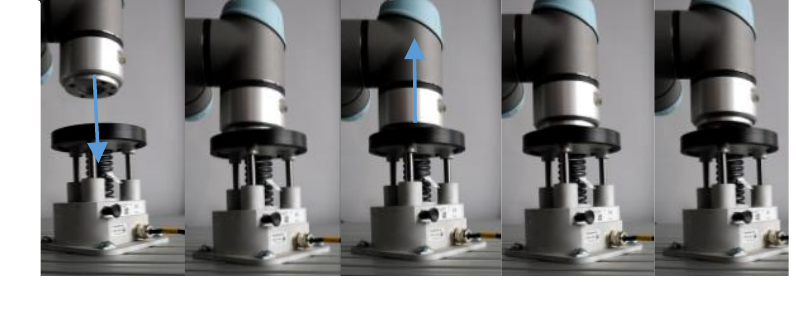}}%
    \put(0.04082224,0.07356398){\rotatebox{89.876603}{\makebox(0,0)[lt]{\lineheight{1.25}\smash{\begin{tabular}[t]{l}Force \end{tabular}}}}}%
    \put(0,0){\includegraphics[width=\unitlength,page=2]{intro.pdf}}%
    \put(0.50906737,0.00049895){\makebox(0,0)[lt]{\lineheight{1.25}\smash{\begin{tabular}[t]{l}Time\end{tabular}}}}%
    \put(0.76834206,0.0087108){\makebox(0,0)[lt]{\lineheight{1.25}\smash{\begin{tabular}[t]{l}$<<$\SI{0.5}{s}\end{tabular}}}}%
  \end{picture}%
\endgroup%

	\vspace*{-2mm}
	\caption{Experimental risk assessment of a constrained contact with integrated robot collision reaction.}
	\label{fig:intro}
	\vspace{-5mm}
\end{figure}

With the advent of modern collaborative robots, the technical specification ISO/TS 15066:2016(E) (TS) \cite{ISO_TS} was published. It shall guide the robot practitioner to ensure human safety in human-robot interaction (HRI) by proposing measures to reduce the risk of human injury. In the TS, the collaboration mode that allows for direct physical interaction between the human and the robot is called Power and Force Limiting (PFL) \cite{ISO_TS, Yamada_1997}. It requires the robot to sense contact and react safely, which can be realized with different well-established methods \cite{haddadin2008collision, deluca_2006, Yamada_1997}, which are partially already embedded in commercially available robots, e.g.,  the Universal Robot's retraction reaction behavior depicted in Fig. \ref{fig:intro}. Essential to PFL is a risk analysis that considers the severity of harm caused by potential collisions. However, defining the severity of harm in such contact situations is challenging because there is a lack of verified tools and injury data
to adequately estimate potential injuries in HRI \cite{Yamada_1997, Haddadin_2012}. Therefore, the TS provides guidance on defining injury severity by establishing thresholds for safe contact forces or pressures during physical HRI based on human pain onset studies \cite{Yamada_1997, Behrens_2019, Melia_2019}. Simplified contact force models and measurement devices based on spring-damper systems as depicted in Fig. \ref{fig:intro} are provided to estimate the force experienced in a contact scenario \cite{ISO_TS, ISO_DIS}. The first step in properly performing a risk assessment for HRI is to translate the real potential hazard into an abstracted form that is consistent with the TS terminology. 
In the context of PFL, the TS provides definitions for the respective contact scenarios. Unfortunately, the contact scenario definitions are deficient and inconsistent, which can lead to different interpretations as in \cite{TUV_2019, Svarny_2021, Kirschner_2021_ra}, for example. 

In this paper, we investigate the different interpretations of the contact scenario definitions in the TS and the corresponding implications on risk assessment. We experimentally validate whether the interpretations lead to TS-compliant contact forces, using four commercial collaborative robots. Finally, we propose concrete improvements to simplify risk assessment and enhance the current standardisation.

This paper is organized as follows. Sec. \ref{sec:contsec} gives an overview of contacts in physical HRI and related definitions. In Sec. \ref{sec:isodef}, we give an overview of contact definitions in the TS and the resulting issues. In Sec. \ref{sec:riskassessment}, we describe the risk assessment process and challenges that users often face. For a KUKA LWR iiwa 14, TM5 700, UR10e, and FE robot, the TS-conformity of suggested velocities from the risk assessment are experimentally validated in Sec. \ref{sec:results}. Then, specific improvements for the risk assessment are proposed in section \ref{sec:improve}. Finally, section \ref{sec:con} concludes the paper.

\section{Definitions and scenarios for contact in HRI}
\label{sec:contsec}



We refer to the following definitions to describe human-robot contact based on \cite{Haddadin_2016}.

\begin{figure*}
    \centering
    \includegraphics[width=1\linewidth]{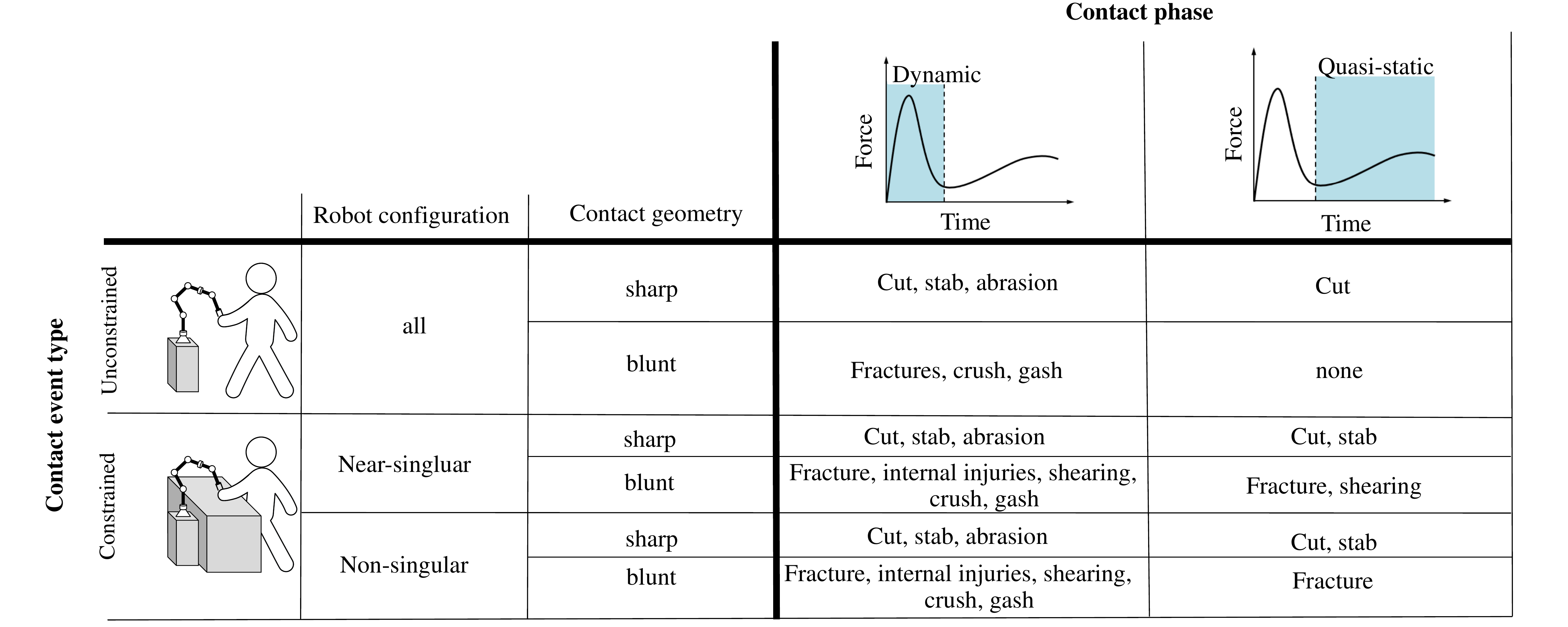}
    \vspace*{-5mm}
    \caption{Fundamental subclasses of contact scenarios and corresponding possible injury scenarios with that should be considered in risk assessments and therefore standardization following \cite{Haddadin_2016}.}
    \label{fig:contact types}
    \vspace*{-5mm}
\end{figure*}

\begin{Theorem}\textbf{HRI. Contact Event Type:}
\label{Def. HRI.1.}
A description of the physical motion constraints to the human body part prior to and during a contact which is independent of time.
\end{Theorem}

\begin{Theorem}\textbf{HRI. Constrained Contact:}
\label{Def. HRI.2.}
A contact event type, where the human body part can not recoil.
\end{Theorem}

\begin{Theorem} \textbf{HRI. Unconstrained Contact:}
\label{Def. HRI.3.}
A contact event type, where the human body part can freely recoil.
\end{Theorem}

\begin{Theorem} \textbf{HRI. Contact Force Phase:}
\label{Def. HRI.4.}
Distinct changes in the contact force profile to distinguish between dynamic and quasi-static contact.
\end{Theorem}

\begin{Theorem} \textbf{HRI. Dynamic Contact Force:}
\label{Def. HRI.5.}
An impact force within the first short impact phase (Phase I) which is mainly defined by the robot and human dynamics. 
\end{Theorem}

\begin{Theorem} \textbf{HRI. Quasi-Static Contact Force:}
\label{Def. HRI.6.}
An impact force remaining in contact force phase II after the dynamic contact force phase and leading to either a pushing scenario (unconstrained contact type) or a clamping scenario (constrained contact type).  
\end{Theorem}

\begin{Theorem} \textbf{HRI. Contact Scenario:}
\label{Def. HRI.7.}
A combination of contact event type, e.g., constrained collision and the contact duration e.g. dynamic contact.
\end{Theorem}








Depending on the contact scenario different possible injuries may occur \cite{Haddadin_2016}; c.f. Fig. \ref{fig:contact types}. The severity of these injuries ranges from skin abrasions and cuts to lacerations and fractures to contusions and puncture wounds. Most importantly, the potential severity of the injury caused is highly dependent on the duration of the contact as well as the person's movement constraints during the contact situation. While a constrained, dynamic contact situation with a robot in a non-singular and sharp geometry may only cause abrasions, it will most likely result in a cut or sting if it persists in the quasi-static phase. While this constrained quasi-static contact with a sharp geometry may even result in a stab to a human body part, it will cause a less severe cut if the type of contact event is not constrained. For an appropriate risk assessment according to DIN EN ISO 12100:2010, the type of possible injury is of great importance \cite{ISO_12100}. A practitioner can observe contact scenarios of different contact event types and during different contact force phases in practical application. However, for an adequate risk assessment, the same practitioner should be able to translate his observations into the terminology of the norms and standards. Because of this translation between practical observation and theoretical analysis of risks in HRI, a uniformly defined terminology of norms and standards is central to the success of safe HRI.

\section{ISO/TS 15066 Definitions human-robot contact}
\label{sec:isodef}

\begin{figure*}
    \centering
    \includegraphics[width=1\linewidth]{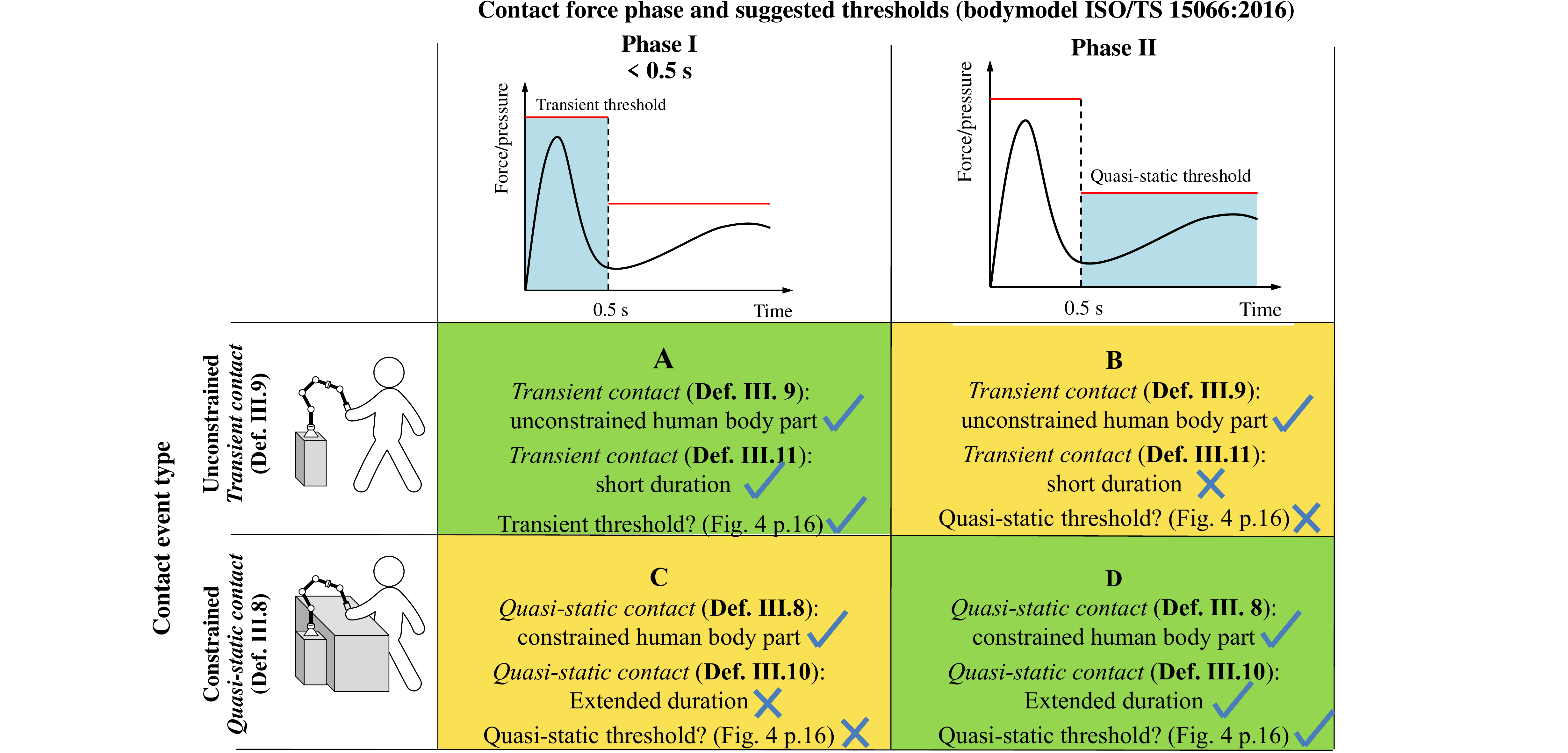}
    \caption{Structure of contact scenarios that are created by the distinction between contacts in TS. The green fields show a consistency of the terminology for the contact type description throughout the TS while the yellow fields include mismatches.}
    \label{fig:prob}
    \vspace*{-5mm}
\end{figure*}

Considering the contact situations depicted by Fig. \ref{fig:contact types}, TS refers only to blunt contact geometries and non-singular robot configurations. It states that sharp contact or contact in a near-singular configuration should be avoided in all cases. As a result, TS proposes the distinction between only two different types of contact scenarios in the beginning of the document:

\begin{Theorem}
\label{Def. TS.1.}
\textbf{TS. Quasi-Static Contact (\cite{ISO_TS}, p.2):}
A contact scenario with a human body part which is constrained in its motion. 
\end{Theorem}

\begin{Theorem}

\textbf{TS. Transient Contact (\cite{ISO_TS}, p.2):}
A contact scenario where a human body part is unconstrained and can recoil at contact with the robot. 
\label{Def. TS.2.}
\end{Theorem}

These two definitions refer to the contact event type (\autoref{Def. HRI.1.}). In order to define force thresholds that are acceptable to the human during contact in constrained or unconstrained contact events, TS distinguishes between the contact force phase by duration of the contact force and the contact event type. In this framework, TS redefines the quasi-static and transient contact types as follows.

\begin{Theorem}
\textbf{TS. Quasi-Static Contact (\cite{ISO_TS}, p.16):}
\label{Def. TS.3.}
A contact scenario with a human body part which is constrained in its motion and results in a force pressure \textbf{for an extended time}. 
\end{Theorem}

\begin{Theorem}
\textbf{TS. Transient Contact (\cite{ISO_TS}, p.16):}
\label{Def. TS.4.}
A contact scenario where a human body part can recoil unconstrained at contact with thus \textbf{short duration}. 
\end{Theorem}

Definitions ~\autoref{Def. TS.3.} and ~\autoref{Def. TS.4.} reflect the force curve \cite{Yamada_1997, Haddadin_2012} shown in TS \cite{ISO_TS}, p.18. But unfortunately, these two definitions mix up the contact type (\autoref{Def. HRI.1.}, which defines constraint and unconstrained motion) and contact force phases (\autoref{Def. HRI.4.}, which explains short and long contact phases); however these are inherently different.

\begin{figure}[thpb]
	\centering
	\includegraphics[scale=0.45]{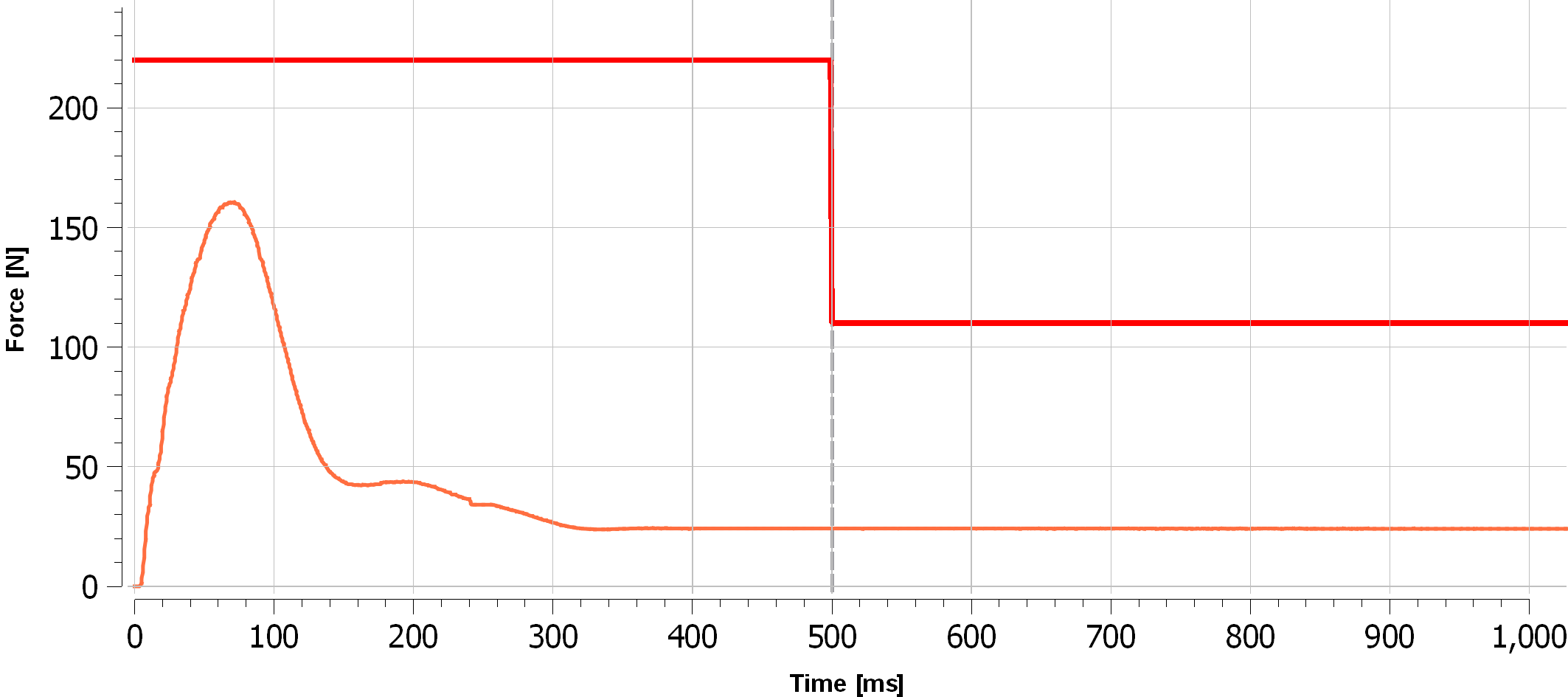}
	\vspace{-3 mm}
	\caption{Exemplary force curve (orange) and comparison against thresholds (red line) for a transient (left) and quasi-static contact force phase (right) supplied by a PRMS device for contact force estimation based on ISO/TS 15066:2016(E) between robot and human belly. }
	\label{fig:force curve PRMS}
	\vspace*{-5mm}
\end{figure}

The principle of the force curve and the corresponding contact force phases are applied to contact force measuring devices, such as the PRMS device; c.f. Fig. \ref{fig:force curve PRMS}. These devices use an automatic evaluation to compare the contact forces occurring in the dynamic and quasi-static phases with the proposed threshold values. These proposed thresholds refer to the \textit{transient contact force} for contacts in the dynamic phase and the \textit{quasi-static contact force} for contacts in the quasi-static phase without any reference to the actual contact event type (unconstrained or constrained). 

This results in the inconsistency depicted in Fig. \ref{fig:prob}. While the TS definitions ~\autoref{Def. TS.3.} and ~\autoref{Def. TS.4.} clearly and consistently define the contact types highlighted in green, the yellow contact types show an inconsistency in the use of terminology that leaves room for interpretations such as: 

\emph{"Quasi-static contact lasts for longer than 0.5 seconds per definition and the human is trapped between the robot and the
surroundings or parts of the robot. In contrast, transient contact lasts a maximum of 0.5 seconds and the affected person
can withdraw or step out of the way after contact ..."} \cite{TUV_2019}, p.17, published by T\"UV AUSTRIA Group.

This interpretation suggests that any contact in which the human can withdraw after a maximum contact time of \SI{0.5}{s} can be considered a \textit{transient contact}, regardless of the nature of the contact event type. While this may be true for contact scenario A, it is not true for scenario C (see Fig. \ref{fig:prob}). Based on this description, the short constraint contact between human and a collaborative robot should be considered as transient since the robot may detect the collision and retracts quickly; however it is quasi-static according to \autoref{Def. TS.1.} definition because the human can't recoil, but still not meeting the conditions of \autoref{Def. TS.3.} and \autoref{Def. TS.4.}

Contact scenario C, the dynamic but constrained contact scenario, is of particular interest for such misinterpretations. Due to the robot's collision response embedded in almost all collaborative robots nowadays, the robot usually backs away or retreats from the human body part. This results in a very short and dynamic contact that does not lead to a quasi-static scenario (clamping) and the human can retreat after the contact. Using TS, this contact scenario can be interpreted as a quasi-static contact (~\autoref{Def. TS.1.}) on the one hand, but on the other hand it does not meet the requirements for a quasi-static contact according to ~\autoref{Def. TS.1.}, since the contact does not exert force over a longer period of time.


\begin{figure*}
    \centering
    \includegraphics[width=1\linewidth]{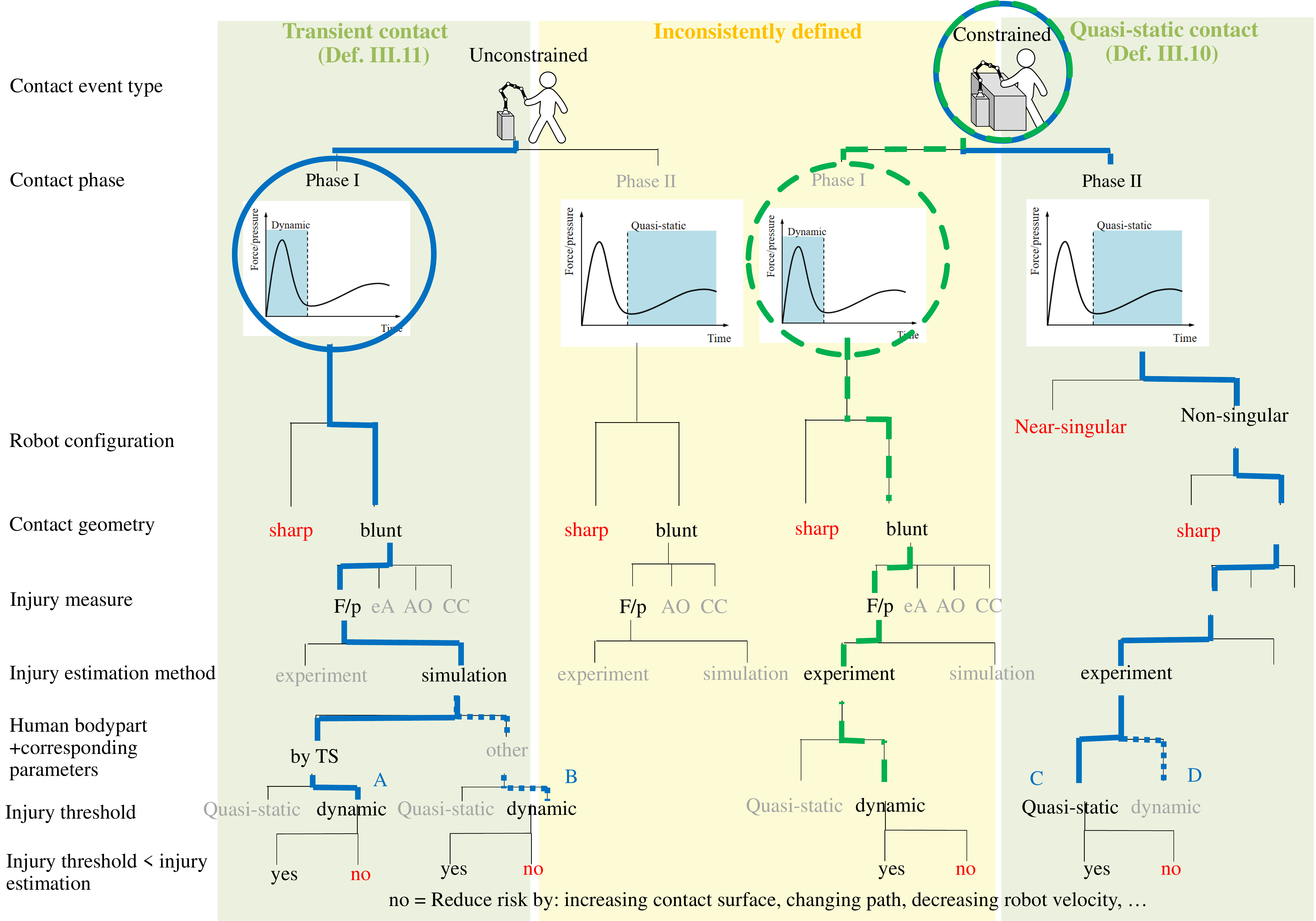}
    \vspace*{-5mm}
    \caption{Decision tree for conducting a risk assessment based on the fundamental contact subclasses \cite{Haddadin_2016}, p. 1846 and adapted for applying ISO/TS 15066:2016(E) and the corresponding test devices for contact in physical HRI. Written in red are criteria immediately resulting in a required risk reduction according to TS, written in grey are options not directly supported by TS. The green area describes the consistently defined \emph{transient} and \emph{quasi-static contact}, while the yellow area describes the inconsistently defined part. For conducting a risk assessment for a constrained but dynamic contact (contact force phase I) based on TS, four interpretations are shown. The main assumption for the contact definition (constrained or dynamic) is circled The blue lines show the interpretations resulting when trying to follow the consistent definitions. The dotted blue line includes own interpretations or deviation from ~\autoref{Def. TS.3.} and ~\autoref{Def. TS.4.} The decision options resulting in the most realistic risk assessment are represented by a green, dashed line. The possible injury measures are referred to as force or pressure (F/p), energy density (eA), compression criterion (CC), AO-classification (AO).}
    \label{fig:ra}
    \vspace*{-5mm}
\end{figure*}

\section{Risk assessment with different interpretations}
\label{sec:riskassessment}
To implement a safe HRI robot cell, the first step is to understand the potential severity of harm in a contact scenario and to sufficiently limit that harm. 
According to the European Council Machinery Directive \cite{mach_dir_2006} and DIN EN ISO 12100:2010, a risk assessment is required \cite{ISO_12100}. This risk assessment should be based on the state of the art represented by norms, standards, and technical specifications. These documents should guide the risk assessment process to ensure best practices for employee safety. To provide an overview of the key steps, Fig. \ref{fig:ra} presents a decision tree for risk assessment focused on the TS and enriched by the work in \cite{Haddadin_2016}. 






The risk assessment starts with defining the contact event type (unconstrained or constrained), followed by the contact force phase (Phase I or phase II). For a constrained and quasi-static contact, the robot configuration is distinguished between near-singular and non-singular. The contact geometry (sharp or blunt) and the injury measure need to be defined to evaluate an potential injury. Following TS, the injury measures of force and pressure (F/p) should be applied based on the data obtained for the onset of pain in humans. The potential occurrence of injuries or the resulting F/p in a contact situation should be estimated by experiment or simulation. In particular, simulation requires a model of the impact dynamics of humans and robots. TS provides a model for determining \emph{transient} contact forces including a list of effective mass and stiffness to be considered for human body parts defined as

 \begin{equation}
    F_\mathrm{col} = v_{rel}\sqrt{\mu k}  \, ,
    \label{eq:ISO}
\end{equation}
 where the human body part stiffness is denoted as $k$, the relative velocity between human and robot as $v_\mathrm{rel}$, and the effective mass in the contact scenario according to \cite{ISO_TS} as
 \begin{equation}
    \mu = \left(\frac{1}{m_\mathrm{r}}+\frac{1}{m_\mathrm{h}}\right) ^{-1} \, ,
    \label{eq:ISO}
\end{equation}
  where $m_\mathrm{h}$ is the human body part mass and
 \begin{equation}
    m_\mathrm{r} = M/2 + m_\mathrm{L} \, ,
 \label{eq:mr}
\end{equation}
 the robot mass. The mass of the robot is defined as an approximation by the total mass of all moving links, denoted $M$, and the load $m_\mathrm{L}$ \cite{ISO_TS}. While TS proposes this simplified and inaccurate model \cite{Kirschner_2021_eff}, an alternative derivation method of the robot effective mass is offered by \cite{Khatib_1995}, based on the robot dynamics. According to \cite{Khatib_1995}, the robot mass perceived at the point of contact in the Cartesian unit direction of impact $\vu \in \mathbb{R}^{3}$ becomes 
\begin{equation}
    m_\mathrm{r,u} = \left ( \vu^\tp \vLambda_\nu^{-1}(\vq) \vu \right )^{-1} \, ,
\label{eq:refl_mass}
\end{equation}
where $\vLambda_\nu^{-1}(\vq)$ is the upper $3 \times 3$ matrix of the robot Cartesian mass matrix inverse
\begin{equation}
\vLambda^{-1}(\vq) = \vJ(\vq) \vM(\vq)^{-1} \vJ(\vq)^\tp \, ,
\label{eq:cart_mass}
\end{equation}
with $\vJ(\vq) \in \mathbb{R}^{n \times m}$ being the Jacobian matrix at the point of contact.

The resulting estimated injury level is compared to the injury threshold provided by TS for quasi-static or dynamic contact. Finally, the contact scenario is classified as safe (yes) or not safe (no) based on the comparison between the estimated injury severity and the injury threshold. If the injury threshold is not met (no), a risk reduction measure should be taken, e.g. reducing the robot speed.


To prevent clamping of a human body part, collaborative robots today often apply post-collision strategies as a safety measure after a collision is detected \cite{haddadin2008collision}. Two examples of this are Universal Robot's UR5e and the FE robot, as seen in \cite{Kirschner_2021_ra}. The application of a collision response strategy leads to a dynamic collision even in the case of unconstrained contact events and resembles contact scenario C in Fig. \ref{fig:prob}. In Table \ref{tab:interpretations}, based on Fig. \ref{fig:ra}, the decisions that can be made to perform a risk assessment based on TS for this constrained, dynamic contact scenario are described. $m_\mathrm{r}$ is the robot effective mass based on ISO while $m_\mathrm{r,u}$is the effective mass calculated according to the robot dynamics. According to TS, $m_\mathrm{h}$ is the human body segment in contact with robot (e.g. hand) and $k$ is the stiffness of the body part. Logically one option is to assume that this value could be $\infty$ in case of constrained collision.





From ~\autoref{Def. TS.1.} it can be deduced that the contact scenario is constrained (\emph{quasi-static}, not \emph{transient}), but the contact duration is really short, as in ~\autoref{Def. TS.4.} (not \emph{quasi-static}, \emph{transient}). Considering our definitions for the type of contact event and the contact force phase, the decision path leading to the most realistic risk assessment is motivated by a constrained contact event type in the dynamic contact force phase I, shown as a green dashed line in Fig. \ref{fig:ra}. Nevertheless, to perform a formal risk assessment, it is still necessary to adhere to the state of the art and therefore stay close to the current definitions of TS. Due to the inconsistent terminology of the TS and further interpretations, such as those provided by \cite{TUV_2019}, the following possibilities arise, which are shown in Fig. \ref{fig:ra} highlighted in blue. While the straight blue lines only consider decisions supported by TS, the dashed blue lines deviate from the suggestions in TS.

\begin{table}[t]
	\caption{ISO-interpretations for risk assessment}
	\vspace*{-8mm}
	\label{tab:interpretations}
	\begin{center}
		\begin{tabular}{p{0.01\linewidth}p{0.9\linewidth}}
			\hline
			\toprule
    \textbf{A} & 
    \tabitem short duration, \emph{transient} following ~\autoref{Def. TS.4.} or \cite{TUV_2019}. \\
    & \tabitem injury estimation: transient contact force model.\\
    & \tabitem model parameters: All from  TS including $m_\mathrm{r}$, $m_\mathrm{h}$ = the effective mass of the body segment, which is in contact with robot (e.g. human hand), and $k$.
    \\ 
    \midrule
   \textbf{B1} & \tabitem short duration, \emph{transient} following ~\autoref{Def. TS.4.} or \cite{TUV_2019}. \\
    & \tabitem injury estimation: transient contact force model. \\
    & \tabitem model parameters: $m_\mathrm{r}$, $k$ from TS and $m_\mathrm{h}$ with the assumption of constrained human body ($m_\mathrm{h}\to \infty$).
    \\
    \midrule
    \textbf{B2} & \tabitem short duration, \emph{transient} following ~\autoref{Def. TS.4.} or \cite{TUV_2019}. \\
    & \tabitem injury estimation: transient contact force model. \\
    & \tabitem model parameters: $k$ from TS, $m_\mathrm{h}$ with the assumption of constrained human body ($m_\mathrm{h}\to \infty$ ), and $m_\mathrm{r,u}$ from \cite{Khatib_1995}. 
    \\
    \midrule
    \textbf{C} & \tabitem constrained, \emph{quasi-static} following ~\autoref{Def. TS.1.} (No model in TS)\\
    & \tabitem injury estimation: experimental.\\
    & \tabitem threshold: quasi-static.
    \\
    \midrule
    \textbf{D} & \tabitem constrained, \emph{quasi-static} following ~\autoref{Def. TS.1.} (No model in TS)\\
    & \tabitem injury estimation: experimental.\\
    & \tabitem threshold: transient (short duration). 
    \\

\bottomrule
		\end{tabular}
	\end{center}
	 \vspace*{-10mm}
\end{table}

While interpretations A, B1, and B2 allow a simulation e.g. for planning, C and D require a purely experimental evaluation as no analytical method for force estimation is provided by TS. For A and B1 we use the contact force model provided by TS, which is described as follows. For interpretation A, the human mass and stiffness for the respective body part are applied, as they are suggested in Appendix A 3.4 \cite{ISO_TS}, p. 27. In B1, an infinite human mass and a stiffness $k =$ \SI{75}{N/mm} for the human hand is assumed. For B2 we apply the reflected mass according to \cite{Khatib_1995}. The following evaluates how accurate the contact force simulation can be. While lower contact forces in reality than in simulation allow for an increase in robot speed in the real scenario and a planned HRI application can be more time efficient than expected, higher actual contact forces lead to contact scenarios that are not TS compliant. In addition, we briefly show the consequence of referring to interpretation C or D with respect to practicality in everyday industrial applications.

\section{Results of the risk assessments}
\label{sec:results}





To illustrate the different interpretations A, B1, and B2 with their actual effects on the resulting contact forces in dynamic forced contact scenarios, it is assumed that once the constrained hand and once the back of the human worker comes into contact with a robot flange. For this scenario, we perform a risk assessment and attempt to realize safe velocities using the above interpretations with four different collaborative robots, namely the LWR iiwa 14 from Kuka (LWR), the TM5 700 (TM5) from Techman, the UR10e from Universal Robot, and the robot from Franka Emika (FE). We investigate

\begin{itemize}
    \item[a)] which velocities and force limits for the robot end effector would be allowed considering the interpretation,
    \item[b)] which contact forces actually result during a contact scenario in different location of the robot reference cube.
\end{itemize}
\vspace*{0mm}

The purely experimental evaluation required for interpretations C and D eventually leads to allowable robot velocities in accordance with TS force thresholds. While D allows higher speeds, C will unnecessarily reduce the robot speed. Therefore, the resulting forces are of secondary importance in these interpretations. Nevertheless, the integration of collaborative robots into flexibly changing work cells should be economical. We therefore measure the times required for our experiments and provide an estimate of the cost required for such an evaluation.


\begin{table*}[t]

	\caption{Risk analysis results for ISO-interpretations hand and back}
	\vspace*{-4mm}
	\label{tab:results}
	\begin{center}
		\begin{tabular}{p{0.02\linewidth}p{0.04\linewidth}p{0.04\linewidth}p{0.04\linewidth}p{0.04\linewidth}p{0.04\linewidth}p{0.04\linewidth}p{0.04
		\linewidth}p{0.04\linewidth}p{0.04\linewidth}p{0.04\linewidth}p{0.04\linewidth}p{0.04\linewidth}p{0.04\linewidth}p{0.04\linewidth}p{0.04\linewidth}}
			\hline
			\toprule
			body part &robot & \multicolumn{2}{c}{\parbox{0.08\linewidth}{$m_\mathrm{r}$ [kg]}} & \multicolumn{2}{c}{\parbox{0.08\linewidth}{$\mu$ [kg]}} & $F_\mathrm{max}$ [N]  & \multicolumn{3}{c}{\parbox{0.12\linewidth}{$v_\mathrm{rel,max} $ [m/s]}} & \multicolumn{2}{c}{\parbox{0.08\linewidth}{$F_\mathrm{A}$ [N]}} & \multicolumn{2}{c}{\parbox{0.08\linewidth}{$F_\mathrm{B1}$ [N]}} & \multicolumn{2}{c}{\parbox{0.08\linewidth}{$F_\mathrm{B2}$ [N]}} \\
			 & & (\ref{eq:mr}) & (\ref{eq:refl_mass}) & ISO values & $m_\mathrm{h} \to \infty$  &  & A & B1 & B2 & C-pos. & N-pos. &  C-pos. & N-pos. & C-pos. & N-pos.   \\
			\midrule
    \multirow{4}{*}{hand} &\textbf{UR10e} & 10.87 & n.p. & 0.57 & \multirow{4}{*}{\parbox{0.04\linewidth}{= $m_\mathrm{r}$}} & \multirow{4}{*}{\parbox{0.04\linewidth}{280}} & 1.25 & 0.28 & - & \multirow{4}{*}{\colorbox{red}{\textcolor{white}{$>500$}}} & \multirow{4}{*}{\colorbox{red}{\textcolor{white}{$>500$}}} & \colorbox{red}{\textcolor{white}{316}} & \colorbox{babyblue}{\textcolor{white}{230}} & - & -\\
    & \textbf{FE} & 5.54 & 2.9 & 0.54 & & & 1.29 & 0.40 & 0.55 & & & 
\colorbox{green}{289} & 
\colorbox{babyblue}{\textcolor{white}{228}} & \colorbox{red}{\textcolor{white}{413}} & \colorbox{orange}{347} \\ 
    & \textbf{LWR} & 6.75  & n.p. & 0.55 &  & & 1.27 & 0.36 & - & & & \colorbox{orange}{368} & \colorbox{green}{279} & - & - \\ 
    & \textbf{TM5 } & 11.35 & n.p. & 0.57 & & & 1.25 & 0.28 & - & & & \colorbox{red}{\textcolor{white}{427}} & \colorbox{orange}{292} & -  & - \\ 
    
			\midrule
    \multirow{4}{*}{back} & \textbf{UR10e} & 10.87 & 0.57 & n.p. & \multirow{4}{*}{\parbox{0.04\linewidth}{= $m_\mathrm{r}$}} & \multirow{4}{*}{\parbox{0.04\linewidth}{420}} & 1.25 & 0.28 & - & \colorbox{babyblue}{\textcolor{white}{379}}  & \colorbox{blue}{\textcolor{white}{196}}  & \colorbox{blue}{\textcolor{white}{257}} & \colorbox{blue}{\textcolor{white}{296}} & - & - \\
    & \textbf{FE} & 5.54 & 2.9  & 0.54 & & & 0.63 & 0.40 & 0.55 & \colorbox{babyblue}{\textcolor{white}{300}} & \colorbox{blue}{\textcolor{white}{237}} & \colorbox{blue}{\textcolor{white}{191}} & \colorbox{blue}{\textcolor{white}{147}} & \colorbox{blue}{\textcolor{white}{269}} & \colorbox{blue}{\textcolor{white}{216}} \\ 
    & \textbf{LWR} & 6.75 & n.p. & 0.55 & & & 0.55 & 0.36 & - & \colorbox{orange}{444} & \colorbox{blue}{\textcolor{white}{276}} & \colorbox{blue}{\textcolor{white}{289}} & \colorbox{blue}{\textcolor{white}{237}} & - & - \\ 
    & \textbf{TM5} & 11.35 & n.p. & 0.57 & & & 0.46 & 0.28 & -  & \colorbox{orange}{443} & \colorbox{babyblue}{\textcolor{white}{354}} & \colorbox{blue}{\textcolor{white}{311}} & \colorbox{blue}{\textcolor{white}{246}} & - & - \\

\bottomrule
		\end{tabular}
		\end{center}
		
		\fbox{\begin{tabular}{llllll}
        \textcolor{red}{$\blacksquare$} &  Force underestimation $>$ \SI{100}{N}. & \textcolor{orange}{$\blacksquare$} &  Force underestimation $>$ \SI{10}{N} &
        \textcolor{green}{$\blacksquare$} &  Correct force approximation by $\pm$ \SI{10}{N}  \\
        \textcolor{babyblue}{$\blacksquare$} &  Force overestimation $>$ \SI{10}{N}. & \textcolor{blue}{$\blacksquare$} &  Force overestimation $>$ \SI{10}{N}  \\
        
\end{tabular}}

	\vspace*{-5mm}
\end{table*}

\subsection{Validation of contact forces}
\begin{figure}
    \centering
    \includegraphics[width=1\linewidth]{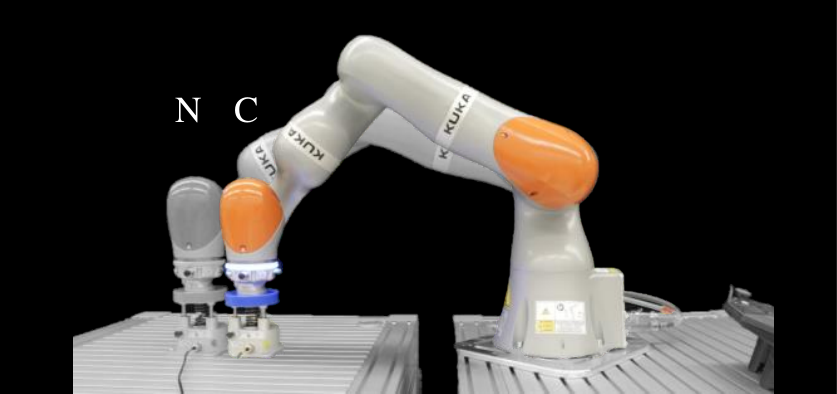}
    \caption{Testsetup to measure forces in a constrained contact scenario with the Kuka LWR iiwa 14 robot and its corresponding central (C) and outer (N) position of the reference cube based on DIN EN ISO 9283:1998 \cite{ISO_9283} }
    \label{fig:setup}
    \vspace*{-3mm}
\end{figure}

To perform the risk assessment, the first thing required for all collaborative robots is the robot mass. Here we face the first challenge. Only for the UR10e robot can we find reliable information from the manufacturer on the joint masses that allow us to estimate the total moving mass of the robot \footnote{https://www.universal-robots.com/articles/ur/application-installation/dh-parameters-for-calculations-of-kinematics-and-dynamics/}. For the FE and the LWR robot, we find estimates from previous researchers in \cite{Gaz_2019} and in \cite{Hennersperger_2017}. For the TM5 robot, we refer to the total mass of the robot and apply (\ref{eq:mr}) since no other data is available. Since the total mass is higher than the effective mass of the actual moving links, we expect an overestimation of the contact forces for the TM5. The TM5 robot also requires an adapter with $m_\mathrm{a} =$ \SI{0.6}{kg}. The resulting robot masses following (\ref{eq:mr}) are listed in Table \ref{tab:results}. Starting from the mass of the robot, the effective mass $\mu$ is derived for interpretation A with the values given in TS Appendix A (hand with $m_\mathrm{h} =$ \SI{0.6}{kg}, $k = $ \SI{75}{N/mm} and back with $m_\mathrm{h} =$ \SI{40}{kg}, $k = $ \SI{35}{N/mm}). For interpretations B an infinite mass of the body part is applied. Finally, using the maximum allowable force $F_\mathrm{max}$ from TS, the velocity limits $v_\mathrm{rel,max}$ are calculated using eq. 2. Table \ref{tab:results} lists the rounding values for all resulting velocities used in the experiments. Using a PRMS\footnote{\url{https://www.pilz.com/download/open/PRMS\_Set\_Supplement\_22225-2EN-04.pdf}} device from Pilz to validate safe contact forces, we measure the maximum occurring contact forces considering the C- and N-positions of the robot reference cube as contact points depicted in Fig. \ref{fig:setup} and the displacement in x-direction listed in Table \ref{tab:refcube} \cite{ISO_9283}. Based on \cite{Kirschner_2021_iros}, we expect the contact forces resulting from the robot speeds in interpretation A to exceed the maximum allowable load on the test apparatus of \SI{500}{N}. Therefore, we do not perform these tests.

We compare whether the measured forces match the expected force calculated using the TS force thresholds. We consider the measured forces that are within $\pm $ \SI{10}{N} to be a reasonable approximation, highlighted in green in Table \ref{tab:results}. Light blue is used for overestimates of more than \SI{10}{N} and dark blue for more than \SI{100}{N}. Orange is used for underestimates of the occurring force by more than \SI{10}{N} and red for more than \SI{100}{N}.

\begin{table} 
	\label{tab:refcube}
	
	\caption{Robot base frame y-coordinate of contact location}
	\vspace*{-2mm}
		\begin{tabular}{p{0.15\linewidth} p{0.15\linewidth} p{0.15\linewidth} p{0.15\linewidth}p{0.15\linewidth}}
			\toprule
			& FE & UR10e & TM5 & LWR  14\\
			\midrule
			$x_\mathrm{N}$ [mm]& $698$ & $926$ & $557$ & $700$ \\

			$x_\mathrm{C}$ [mm]& $498$& $726$ & $357$ & $600$ \\
			
			\bottomrule

		\end{tabular}
		\vspace{-0.5cm}
\end{table}

Finally, for the interpretation A in contact with the human hand, we note that none of the contact forces of the robot at both positions would correspond to the calculated force. Thus, using this velocity would result in a violation of the defined force thresholds in each case. For Interpretation B1, the calculated force provided an adequate estimate of the actual contact force in two situations. In two cases, there is an underestimation and an overestimation below \SI{100}{N}. However, for the UR10e and the TM5 robot in the C-position, the contact force was underestimated by more than \SI{100}{N}. This is particularly surprising since the total mass of the TM5 robot was used in the calculation, not just the mass of the moving links. Therefore, one would expect lower rather than higher force values. For interpretation B2, only the full dynamics model of the FE robot is available, so only this robot is tested. Although the robot's mass is position-dependent and calculated with the best possible state-of-the-art estimate, we observe an underestimation of the actual force by more than \SI{100}{N} in the C-position and less than \SI{100}{N} in the N-position. For the human hand, none of the interpretations lead to acceptable results and each interpretation results in an underestimation of more than \SI{100}{N} for at least one of the considered robots. Therefore, equation (\ref{eq:ISO}) should not be used to estimate dynamic contact forces with the human hand.

If, on the other hand, the human shoulder or back is considered, a strong overestimation of the actual contact forces by more than \SI{100}{N} is observed. The overestimation applies to all test conditions with interpretation B1 as well as B2 and is still valid for interpretation A UR10e, FE, and LWR in N-position. In the C-position, the contact forces of the FE robot and the UR10e robot are overestimated by less than \SI{100}{N}, and in the case of the LWR  14 and the TM5, an underestimation of less than \SI{100}{N} occurs. In summary, in most cases the contact force model leads to an overestimation of the contact forces of the robot with the human shoulder or back, resulting in unjustified robot velocity reduction.


\subsection{Profitability for an experimental risk assessment}

The experiment preparation for each collision position and robot adjustment requires $\approx$ \SI{0.5}{h}. The robot speed adjustment is measured to be approximately \SI{0.05}{h}, depending on the robot. Repeating the experiment three times takes \SI{0.02}{h} on average. To find the most efficient and at the same time the safest robot settings considering TS without prior knowledge of the safe speed, $\approx 6$ trials were required. This results in a time of about \SI{0.87}{h} for risk reduction in a single position, one human body part, and without considering pressure distribution or any trajectory adjustments that may be required. In a real application, multiple contact points may be relevant, the robot trajectory may need to be changed, or even parts of the robot end-effector may need to be readjusted. Assuming three contact points for the evaluation and at least two body parts to be considered, the time required to adjust a system is around \SI{5.22}{h}. This indicates that validation of the force thresholds after even a minor adjustment of the robotic cell can require up to one working day of downtime for the machine and the factory worker or engineer. Since a validation of force thresholds is required every time the robot cell is changed, a safety validation following TS makes the idea of flexible workspaces with HRI cells uneconomical. 

Finally, for a reliable risk assessment in dynamic and constrained contact scenarios, we found the three major challenges:

\begin{itemize}
    \item[1.] Neither of the three interpretations A, B1, or B2 enables a reliable approximation of the location dependent contact forces,
    \item[2.] interpretation C causes unnecessary robot speed reduction, and
    \item[3.] the time requirement of interpretation C and D for a successful risk reduction and validation of this risk is not profitable.
\end{itemize}

\section{Discussion and suggestions for constrained contact risk assessment}
\label{sec:improve}

Overall, we observe large differences in  the allowed robot velocity and the resulting contact forces depending on the interpretations of the TS. This analysis shows the importance of the distinction between the contact event type and the contact force phase as well as the consistency of the definitions for all possible contact scenarios, but especially for the dynamic, constrained contact situations observed in this work. In the draft version ISO/DIS 10218-2:2020-12 (DIS), which is intended to integrate ISO/TS 15066 with the former EN ISO 10218-1, the definitions ~\autoref{Def. TS.1.} and ~\autoref{Def. TS.2.} have been moved to footnotes and additional information has been provided on the experimental derivation of contact forces. While this is a step towards clarifying the issues discussed in this paper, it does not yet adequately define the difference between contact scenarios. Since the translation between real-world scenario and terminology in standards contributes significantly to the safety and use of collaborative robots, the authors recommend revising the definitions chosen in the TS and its successors and eliminating the use of the terms \emph{transient} and \emph{quasi-static} to describe both the contact force phase and the type of contact event. Instead, we propose to use the terms \emph{unconstrained} and \emph{constrained} for the contact event type (cf. \cite{Haddadin_2016}) and \emph{dynamic} and \emph{quasi-static} for the contact force phase. In addition to defining contact scenarios, other researchers report more fundamental definitions in HRI that need to be clarified, such as the terms collaboration, coexistence, or cooperation \cite{Vincentini_2020}. Also these should be clarified by a future version of TS.

\begin{figure}
    \centering
    \includegraphics[width=1\linewidth]{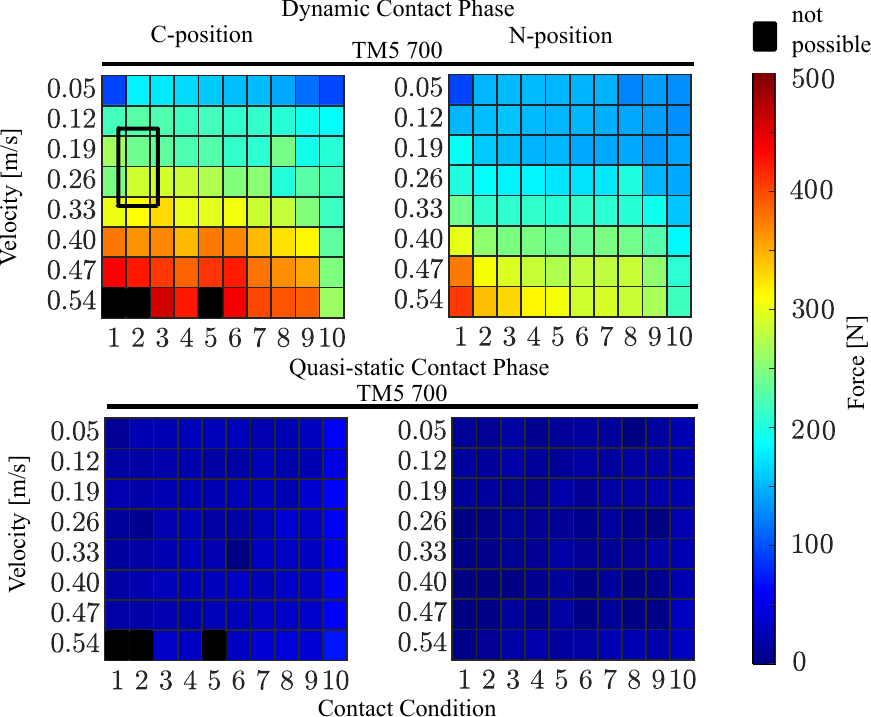}
    \caption{Constrained collision force maps for the Techman TM5 robot and all body location defined by TS. The map shows the contact forces that can be expected when the robot collides at a certain velocity with the body part in either central (C) position or outer (N) position of the robot reference cube. By searching for the force values which comply to the body part's force thresholds the integrator can easily obtain robot velocities which are conform to TS, e.g. for the human hand in dynamic contact. }
    \label{fig:CCFMs}
    \vspace*{-5mm}
\end{figure}

Since the inconsistency in the definition of a dynamic constrained contact scenario may lead to the application of the contact force model for \emph{transient} contact, we considered in particular the results of this assumption. We observe a significant influence of the robot configuration on the contact forces, which the robot mass model of TS (\ref{eq:mr}) cannot account for. Even if we use a more appropriate mass model (\ref{eq:refl_mass}) that takes into account the robot configuration, we still observe strong variations between predicted and real contact forces. Consequently, we can say that the contact force model (\ref{eq:ISO}) is not suitable for estimating dynamic unconstrained contact forces in any of the above interpretations. In the author's experience, a model for dynamic unconstrained contact forces would require information about the robot's contact sensitivity, force thresholds, and braking distances, which depend on the robot configuration, its sensing performance, and the performance of the robot controller, among other factors \cite{Kirschner_2021_iros}. This causes highly complex models for the estimation of constrained contact forces. Nevertheless, the severity of harm in a dynamic constrained contact scenario is potentially high \cite{Haddadin_2016} and requires proper consideration in a risk assessment. Therefore, the authors propose to refrain from model-based analysis of robot contact forces in dynamic, unconstrained contact scenarios. To provide practitioners with a comprehensive guide for risk assessment, the authors first propose a visualization of the required steps through decision trees, as shown in Fig. \ref{fig:ra}. To avoid further problems arising from the lack of a model to estimate the dynamic and quasi-static contact forces, we propose to apply predefined maps of the robot contact forces for different contact points. We proposed \textit{constrained collision force maps (CCFM)} for the quasi-static and dynamic contact force phases \cite{Kirschner_2021_ra, Kirschner_2021_iros}, shown in Fig. \ref{fig:CCFMs} for the TM5 robot. The human contact stiffness and damping parameters proposed by TS are labeled 1-10 following \cite{Kirschner_2021_iros}. The application of these maps is provided as follows. We assume a given robot trajectory, which can lead to a constrained dynamic contact scenario with the human hand. The maximum allowable force of \SI{280}{N} is given by TS. The contact point is located near the center of the robot reference cube. Using the map, the practitioner can now determine the range of robot velocities where the force threshold of TS is likely to be reached, based on the color coding. He/she can begin the experimental risk assessment with a velocity in this range, requiring only a few iterations to find the optimal solution. This can shorten the duration of the experimental validation of the contact forces. In addition, even without experimental validation, the probability that the velocity selected based on the experimentally derived CCFMs will match the TS is higher than using any interpretation of the force model. Therefore, the next step towards simplified robot integration is to embed this knowledge into the robot controller.

In this work, the interpretations of the existing TS for collaborative robots are discussed, accepting force and pressure thresholds to describe human pain sensation. However, the drawback of using force and pressure as thresholds is evident in this work. Force and pressure at a contact point are responses to a specific set of robot and human parameters and are therefore difficult to estimate and control. Thus, research should focus on describing the hazard potential of a contact in HRI using the directly associated robot parameters instead, as proposed in \cite{Haddadin_2012}. Thresholds for human pain perception can also be controversial, as they do not reflect the actual potential for injury as severity of harm, which is required for risk assessment \cite{Rosenstrauch_2017, Haddadin_2012, ISO_12100}. Future work will discuss these topics in more detail.

\section{Conclusion}
\label{sec:con}
In this paper, we showed how inconsistent terminology in the ISO/TS 15066:2016(E) (TS) can affect the risk assessment for dynamic, constrained contact scenarios. Based on the inconsistent definitions of the terms \emph{transient} and \emph{quasi-static} in the context of a contact event type as well as the contact force phase, we identified different interpretations. The implications of these interpretations were analyzed in exemplary risk assessments with a Franka Emika robot, Techman TM5 700, Kuka LWR iiwa 14, and Universal Robot's UR10e. Three of the possible interpretations employ the contact force model provided by TS. Here, we compared the theoretical contact forces with real ones measured by a PRMS device. The forces differ by more than \SI{100}{N}. We demonstrated that the safety evaluation driven by purely experimental contact force evaluation is very time-consuming. To improve the risk assessment procedure, we first provided terms that consistently describe the contact force phase and the type of contact event. We also provided a decision tree that guides the risk assessment process. Finally, we provided constrained collision force maps to speed up the experimental process of risk mitigation in HRI. Based on this work, we propose to change the ISO/TS 15066:2016(E) terminology in future versions to avoid misleading interpretations and avoid the foreseeable misuse of the TS. Furthermore, we suggest to include our proposed methods for simplified risk assessment. 

\addtolength{\textheight}{-12cm}   





\section*{ACKNOWLEDGMENT}
The authors would like to thank the Bavarian State Ministry for Econmonic Affairs, Regional Development and Energy (StMWi) for financial support as part of the project SafeRoBAY (grant number: DIK0203/01). We gratefully acknowledge the funding of the Lighthouse Initiative Geriatronics by StMWi Bayern (Project X, grant no. IUK-1807-0007// IUK582/001) and LongLeif GaPa gGmbH (Project Y), funding of the Lighthouse Initiative KI.FABRIK Bayern by StMWi Bayern (KI.FABRIK Bayern Phase 1: Aufbau Infrastruktur and KI.Fabrik Bayern Forschungs- und Entwicklungsprojekt, grant no. DIK0249). This work was also supported by the European Union's Horizon 2020 research and innovation programme as part of the project I.AM. (grant no. 871899) and Darko (grant no. 101017274). Please note that S. Haddadin has a potential conflict of interest as shareholder of Franka Emika GmbH.



\bibliographystyle{IEEEtran} 
\bibliography{bib}

\end{document}